\documentclass[draftcls]{IEEEtran}
\usepackage{hyperref}       
\usepackage{url}            
\usepackage{booktabs}       
\usepackage{amsfonts}       
\usepackage{nicefrac}       
\usepackage{microtype}      
\usepackage{calc,amsfonts,amssymb,amsmath,bm,url,color,theorem, cite}
\usepackage[inline]{enumitem}
\usepackage{multirow}
\usepackage{wrapfig}

\usepackage{graphicx}
\usepackage{epstopdf}
\usepackage{algorithm}
\usepackage{algorithmic}
\usepackage{overpic}
\usepackage{subcaption}

\newcommand{\st}{\text{s.t.}}

\newcommand{\one}{\boldsymbol{1}}
\newcommand{\zero}{\boldsymbol{0}}

\newcommand{\X}{\boldsymbol{X}}
\newcommand{\Y}{\boldsymbol{Y}}
\newcommand{\W}{\boldsymbol{W}}
\renewcommand{\d}{\boldsymbol{d}}
\renewcommand{\H}{\boldsymbol{H}}
\newcommand{\D}{\boldsymbol{D}}
\newcommand{\A}{\boldsymbol{A}}
\newcommand{\B}{\boldsymbol{B}}

\newcommand{\U}{\boldsymbol{U}}

\renewcommand{\S}{\boldsymbol{S}}

\newcommand{\eye}{\boldsymbol{I}}
\newcommand{\x}{\boldsymbol{x}}
\newcommand{\s}{\boldsymbol{s}}

\newcommand{\y}{\boldsymbol{y}}
\newcommand{\z}{\boldsymbol{z}}

\renewcommand{\u}{\boldsymbol{u}}
\renewcommand{\a}{\boldsymbol{a}}
\renewcommand{\b}{\boldsymbol{b}}

\renewcommand{\k}{\boldsymbol{k}}
\newcommand{\Q}{\boldsymbol{Q}}

\newcommand{\R}{\mathbb{R}}

\newcommand{\ttt}{\textup{\sf T}}

\newtheorem{Fact}{Fact}
\newtheorem{Lemma}{Lemma}
\newtheorem{Prop}{Proposition}
\newtheorem{Theorem}{Theorem}
\newtheorem{Def}{Definition}
\newtheorem{Corollary}{Corollary}

\theorembodyfont{\rmfamily}

\newtheorem{Remark}{Remark}

\definecolor{orange}{RGB}{255,107,0}

\begin{document}

%

%

%
\title{Learning Nonlinear Mixtures: Identifiability and Algorithm}
\author{\IEEEauthorblockN{Bo Yang \IEEEauthorrefmark{1}, 
Xiao Fu \IEEEauthorrefmark{2}, Nicholas D. Sidiropoulos \IEEEauthorrefmark{3}, Kejun Huang \IEEEauthorrefmark{4}}

\IEEEauthorblockA{\IEEEauthorrefmark{1}Department of Electrical and Computer Engineering \\ 
University of Minnesota, Minneapolis, MN 55455, USA\\
Email: yang4173@umn.edu\\}
\IEEEauthorblockA{\IEEEauthorrefmark{2}School of Electrical Engineering and Computer Science\\
Oregon State University, Corvallis, OR 97330, USA \\
Email: xiao.fu@oregonstate.edu\\}
\IEEEauthorblockA{\IEEEauthorrefmark{3}Department of Electrical and Computer Engineering \\
University of Virginia, Charlottesville, VA 22904, USA\\
Email: nikos@virginia.edu\\}
\IEEEauthorblockA{\IEEEauthorrefmark{4}Department of Computer and Information Science and Engineering \\
University of Florida, Gainesville, FL 32611, USA\\
Email: kejun.huang@ufl.edu}
}
%
\maketitle

\begin{abstract}
  Linear mixture models have proven very useful in a plethora of applications, e.g., topic modeling, clustering, and source separation. 
  As a critical aspect of the linear mixture models, identifiability of the model parameters is well-studied, under frameworks such as independent component analysis and constrained matrix factorization. Nevertheless, when the linear mixtures are distorted by \emph{an unknown} nonlinear functions -- which is well-motivated and more realistic in many cases  --  the identifiability issues are much less studied. This work proposes an identification criterion for a nonlinear mixture model that is well grounded in many real-world applications, and offers identifiability guarantees. A practical implementation based on a judiciously designed neural network is proposed to realize the criterion, and an effective learning algorithm is proposed. Numerical results on synthetic and real-data corroborate effectiveness of the proposed method.
\end{abstract}

\section{Introduction}

Linear mixture models (LMMs) have found numerous applications in machine learning and signal processing, e.g., topic mining, clustering, and source separation. When LMM is used for applications that are essentially parameter estimation (e.g., topic mining and community detection), it is critical to ensure that the generative model is uniquely identifiable. This is also found critical in many data mining problems \cite{huang2016anchor,mao2017mixed}, {as interpretability naturally relates to model uniqueness}. However, LMM is not identifiable in general -- even in the best case without noise: an LMM boils down to a matrix factorization (MF) model that is known to be unidentifiable, unless additional constraints on the factors are imposed.


Identifiability research for LMMs has a long and fruitful history in the confluence of machine learning, statistics, and signal processing. The arguably most notable line of work is independent component analysis (ICA) \cite{comon1994independent, hyvarinen1999survey}, which is motivated by speech source separation. Statistical independence of latent parameters (i.e., different sources) is utilized to establish identifiability. LMM unmixing with correlated latent parameters has also been extensively studied, e.g., in the context of \textit{nonnegative matrix factorization} (NMF) \cite{donoho2004does, laurberg2008theorems, anandkumar2012spectral, huang2014non, fu2015blind, lin2015identifiability, mao2017mixed, ma2014signal}, bounded component analysis (BCA) \cite{cruces2010bounded}, and some other types of constrained MF models \cite{georgiev2005sparse, barak2015dictionary}.

Despite the relatively good understanding to the identifiability issues of different LMMs, the model is considered over-simplified in many applications.
In many cases the observed data cannot be assumed to be approximately linear mixtures of some basis vectors, since nonlinear distortions exist due to a lot of reasons---e.g., multiplicative noise, clipping effect of sensors, and quantization, just to name a few.
A natural question then is: under a reasonable \textit{nonlinear mixture model}, can we identify the latent parameters of interest uniquely? 

This question turns out to be highly nontrivial: most of the analytical tools in the linear mixture case do not apply.
One exception is statistical independence of random variables, which is not affected by nonlinear distortion. Based on this observation, many works \cite{taleb1999source, achard2005identifiability, hyvarinen2016unsupervised, hyvarinen2017nonlinear} tackle nonlinear mixture model identification from a nonlinear ICA viewpoint. This line of work is very elegant, but it only answers our research question partially. Furthermore, statistical independence is considered restrictive, which is one of the main motivations for the extensive study of correlated components / sources as mentioned above.


\paragraph{Contributions.} 
In this work, we study the nonlinear mixture model learning problem, under a new setting that is rather different from ICA.
Specifically, we study a nonlinear mixture model where the observed data vectors are convex combinations of a set of basis vectors followed by a nonlinear distortion. As mentioned, this kind of mixture model finds applications in MRI sensing, hyperspectral imaging, and statistical learning -- and thus is very well-motivated.
Our detailed contributions are
\begin{enumerate}	
	\item \textbf{Identification criterion} We propose a model identification criterion for the considered problem and provide sufficient conditions under which the model is identifiable. Our proof is a novel integration of functional equations \cite{efthimiou2010introduction, kuczma2009introduction} and a generalization of LMM identifiability results, which is a fortuitous union that fits the considered nonlinear model well;
	\item {\bf Neural network-based implementation} We propose a neural network based formulation to implement the proposed criterion. The employed neural network is judiciously designed so that some specific constraints specified by the proposed identification criterion can be satisfied;
	\item {\bf Numerical validation} We reformulate the criterion to an easy-to-implement form  and employ a trust region algorithm for solving the problem efficiently. We also tested the algorithm on both synthetic and real data to show effectiveness of the approach.
\end{enumerate}
Another salient feature of our method is that it turns the \emph{unsupervised} parameter estimation problem into a \emph{supervised} regression problem, which requires little new algorithmic design -- see Section \ref{ch:learning} for more information. 

\paragraph{Notation.} 
Bold capital letters represent matrices, while bold lowercase letters denote vectors, which are assumed to be column vectors, unless transposed with $(.)^\ttt$. Plain lowercase letters denote scalars. $\X$ and $\x$ refer to the observed data, and $\A$, $\a_i$, $\S$, $\s_i$ refer to the underlying latent parameters. {Symbol} 
$\boldsymbol{\phi}$ denotes the unknown nonlinear function in data generation, and $\boldsymbol{f}$ denotes the learning function, which tries to counteract the nonlinear effects in $\boldsymbol{\phi}$. Symbol $\Y$ represents the data transformed by the learning function $\boldsymbol{f}$, i.e. $\Y = \boldsymbol{f}(\X)$, and $\boldsymbol{k}$ denotes the composite function of $\boldsymbol{f}$ and $\boldsymbol{\phi}$. Symbol $[N]$ denotes the set of integers $\{1, \cdots, N\}$. The vector-valued functions we consider in this work are all element-wise, and we use the notation $\boldsymbol{f} = [f_1,\cdots, f_M]^\ttt$ to mean that $[\boldsymbol{f}(\x)](i) = f_i(\x(i))$ for $\x\in \R^{M}$ and $i\in [M]$. The symbol $\|\cdot\|_0$ denotes the $\ell_0$ norm, i.e. the number of nonzeros, of a vector or matrix. The symbol $\text{cone}(\X)$ denotes the set formed by conical combination of columns of $\X$. {Finally,} $\zero$ ($\one$) denotes a vector (or matrix) of all 0's (1's). 

\section{Preliminaries}
We briefly review existing parameter identification results that are related to this work. Relevant concepts in convex geometry can be found in the appendix.

To facilitate discussion, we use
$
\Delta_M := \left\lbrace \x | \x\in \R^M, ~ \x \geq \zero, ~ \one^\ttt\x=1\right\rbrace
$
to denote the $(M-1)$ probability simplex. 
The LMM is defined as
\begin{align}\label{eq:LMM}
\x_j = \A\s_j,~j \in [N],
\end{align}
where $\A \in \R^{M\times r}$ is often a tall matrix, i.e., $M > r$, and $\s_j \in \Delta_r$. Alternatively, we will also write $\X = \A\S$ by collecting all $\x_j$'s into $\X$, and $\s_j$'s into $\S$.

%

In order to characterize identifiability of \eqref{eq:LMM}, let us introduce the following definition.
\begin{Def}{(Sufficiently scattered, \cite{fu2015blind, huang2016anchor})}\label{as1}
	Let matrix $\S \in \R_+^{r\times N}$, where $\R_+^{r\times N}$ is the nonnegative subset of $\R^{r\times N}$. Matrix $\S$ is said to be sufficiently scattered (SS) if $\text{cone}(\S)$ satisfies:
	\begin{enumerate*}[label={(\alph*)}]
		\item $\mathcal{C} \subseteq \text{cone}(\S)$, where $\mathcal{C}$ is a second order cone:
		$
		\mathcal{C} = \{\x\in \R^r | \one^\ttt\x \geq \sqrt{r-1} \|\x\|_2  \},
		$
		\item $\text{cone}(\S) \subsetneq \text{cone}(\Q)$, for any unitary matrix $\Q \in \R^{r\times r}$ that is not a permutation matrix.
	\end{enumerate*}
\end{Def}
Roughly speaking, this condition requires that the column of $\S$ are spread out on the probability simplex. This condition is in fact fairly relaxed, as discussed in \cite{huang2018learning}.

To recover factors $\A$ and $\S$ from data $\X=[\x_1, \cdots, \x_N]$, the following so-called {\em Volume Minimization} (VolMin, \cite{fu2015blind}) criterion is often employed:
\begin{align}\label{eq:volmin}
\min_{\B \in \R^{M\times r}, \H \in \R^{r\times N}} &~~ \text{Vol}(\B) \quad \nonumber\\
\st &~~ \X = \B\H, \nonumber \\
 &~~\H \geq {\bf 0}, ~ \H^\ttt \one  = \one,
\end{align}
where it is assumed that $r$ is known. The term $\text{Vol}(\B)$ is a measure of the volume of the simplex formed by using columns of $\B$ as vertices, see \cite{boyd2004convex}. This criterion suggests that we want to find $\B$ and $\H$ that satisfy the LMM, and we pick the solution with minimal volume, hence the name VolMin. 

Based on this VolMin criterion, the following theorem established identifiability of model\eqref{eq:LMM}.
\begin{Theorem}{(\cite{fu2015blind})}\label{thm:volmin-ident}
	Let the matrices $\A$ and $\S$ satisfy $\text{rank}(\A) = \text{rank}(\S) = r$. Suppose $\S$ satisfies the SS condition.
	Under the generative model \eqref{eq:LMM}, the VolMin criterion \eqref{eq:volmin} uniquely identifies both $\A$ and $\S$ up to a permutation. Specifically, any optimal solution to \eqref{eq:volmin}  takes the form
	\begin{align*}
	\B = \A \boldsymbol{\Pi}, ~ \H = \boldsymbol{\Pi}^\ttt \S,
	\end{align*}
	where $\boldsymbol{\Pi}$ is a permutation matrix.
\end{Theorem}
A proof of this result can be found in \cite{fu2015blind}. We mention that by Theorem~\ref{thm:volmin-ident}, given that $\S$ satisfies SS, the only remaining indeterminacy is a permutation of the columns (rows) of $\A$ (resp. $\S$), which is unavoidable -- but also inconsequential in most applications.

Several algorithms for dealing with \eqref{eq:volmin} have been developed, and we will use the so-called {\em minimal volume enclosing simplex} (MVES): Given data $\X$ and the rank parameter $r$, the MVES algorithm returns a solution $(\widehat{\B}, \widehat{\H})$ of \eqref{eq:volmin}. We refer readers to \cite{chan2009convex} for more on MVES due to page limitations.

\section{The nonlinear mixture model}\label{sec:main}
\subsection{The model}
We introduce a new data model to handle nonlinear effects in various applications. Specifically, the data model is 
\begin{align}\label{eq:non-LMM}
{\x}_j = \boldsymbol{\phi}(\A\s_j), ~ j \in [N],
\end{align}
where $\A\in \R^{M\times r}$ satisfies $\A \geq \zero$, and $\s_j \in \Delta_r, ~\forall j \in [N]$. The function $\boldsymbol{\phi}$ is a nonlinear mapping $\boldsymbol{\phi}: \R^{M} \rightarrow \R^{M}$, and we consider \emph{element-wise} nonlinearity, i.e., $\boldsymbol{\phi} = [\phi_1, \phi_2, \cdots, \phi_M]^\ttt$, so that
\begin{align}
\boldsymbol{\phi}(\x) = [\phi_1(\x(1)), \cdots, \phi_M(\x(M))]^\ttt,
\end{align}
where $\x = [\x(1), \cdots, \x(M)]^\ttt$. For notational brevity, we use the shorthand ${\X }= \boldsymbol{\phi}(\A\S)$ to denote \eqref{eq:non-LMM}, where it should be noted that the $\boldsymbol{\phi}$ is applied on each \emph{column} of $\A\S$. 

Model \eqref{eq:non-LMM} is well motivated. It can be viewed as a generalization of \eqref{eq:LMM}, which is used in various applications. 
In hyperspectral unmixing (HU), each $\x_j$ is a hyperspectral pixel, each column of $\A$ represents the frequency signature of a certain material (e.g. soil, vegetation, water), and each $\s_j$ denotes the proportion of materials in that pixel $\x_j$, see e.g. \cite{bioucas2012hyperspectral, ma2014signal}. {In magnetic resonance imaging (MRI), LMM is used due to the so called ``partial volume effect'' \cite{chan2008convex, wang2006blind, santago1995statistical}, which gives rise to  the condition $\s_j \in \Delta_r$.} 
Both these applications are of great importance in their respective research fields, where considerable work has been done based on \eqref{eq:LMM}. Yet, it is widely recognized that in many real world scenarios, the LMM in \eqref{eq:LMM} is oversimplified, see 
\cite{dobigeon2014nonlinear}. For example, in HU and MRI, the measurements $\x_j$'s are obtained by sensors, which have inherent nonlinearity due to physical limitations of the measuring devices. By explicitly modeling this nonlinearity, we expect methods that are based on \eqref{eq:non-LMM} to give improved results in these tasks. 


For faithful modeling purpose, \eqref{eq:non-LMM} adds the mapping $\boldsymbol{\phi}$ to \eqref{eq:LMM}, which renders \eqref{eq:non-LMM} flexible in covering many important applications, as discussed above. However, it is clear that the additional $\boldsymbol{\phi}$ brings considerable complication in recovering $\A$ and $\S$. Before pursuing a general result, let us make some simple observations. First, for many nonlinear $\boldsymbol{\phi}$, it is not possible to recover $\A$ and $\S$, e.g., $\boldsymbol{\phi}(\x) = \zero, ~\forall ~\x$. Hence one of the tasks is to impose on $\boldsymbol{\phi}$ reasonable and practical conditions, under which recovery is possible. Second, if $\boldsymbol{\phi}$ is linear, by the element-wise assumption, we have $\X = \D\A\S$, where $\D$ is a diagonal matrix. From here, we can see that there are scaling ambiguities on the rows of $\A$, even for the simplest $\boldsymbol{\phi}$. In light of this, a crucial question about model \eqref{eq:non-LMM} is which parts (or aspects) of $\A$ and $\S$ can be identified, and to what extend?

\subsection{Functional equations on a simplex}\label{ch:theory} 
We aim at identifying parameters from \eqref{eq:non-LMM} in an \emph{unsupervised} fashion. Towards that end, we will try to learn an adjustable function $\boldsymbol{f}$, and denote
\begin{align}\label{eq:training}
\y_j = \boldsymbol{f}(\boldsymbol{\phi}(\A \s_j)), \quad j \in [N].
\end{align}
The remaining question is how to devise a learning method such that the resulting $\boldsymbol{f}$ will `counteract' the nonlinear effect brought by $\boldsymbol{\phi}$. If this can be done, we can then employ methods designed for LMM \eqref{eq:LMM} to separate the latent factors. Towards this goal, we first introduce a technical lemma. 



Consider the following functional equation concerning functions $\psi_1, \cdots, \psi_M$ and variables $\s\in \text{int}~\Delta_r$
\begin{align}\label{eq:dr_1}
\sum_{i=1}^{M}\psi_i(\a^\ttt_i \s) = 1, ~ \forall \s \in \text{int}~\Delta_r,
\end{align}
where $\text{int}~\Delta_r$ denotes the \emph{interior} of $\Delta_r$.
To facilitate presentation, let $\A := [\a_1, \a_2, \cdots, \a_M]^\ttt \in \R^{M\times r}$. 


\begin{Lemma}\label{lemma-dr}
	Suppose \eqref{eq:dr_1} holds, and $M \geq r\geq 3$. Let us further assume that
	\begin{enumerate*}[label=(\alph*)]
		\item  the functions $\psi_1, \cdots, \psi_M$ are twice differentiable, and are all convex (or all concave) in the domain $(0, 1)$; and 
		\item  $\A$ is nonnegative and has two positive columns.
	\end{enumerate*}
	Then the functions $\psi_1, \cdots, \psi_M$ are all affine.
\end{Lemma}
The proof can be found in the appendix.

\subsection{Nonlinear mixture model identification}
To proceed, let us suppose that the learning
function $\boldsymbol{f}: \R^{M} \rightarrow \R^M$ in \eqref{eq:training} is also element-wise, i.e., $\boldsymbol{f} = [f_1, f_2, \cdots, f_M]^\ttt$, where $f_i$'s are univariate functions.
Denote $\k = [k_1, k_2, \cdots, k_M]^\ttt: \R^M \rightarrow \R^M$, where $k_i = f_i \circ {\phi}_i$, and $\circ$ denotes function composition. Let us make the following assumptions about the generative model \eqref{eq:non-LMM}.

\begin{enumerate}[label={(A\arabic*)},ref={(A\arabic*)}]
	\item  \label{as-phi}  The functions $\phi_1, \cdots, \phi_M$ are all invertible, and twice differentiable. 
	\item  \label{as-A} The matrix $\A\in \R^{M\times r}$ in \eqref{eq:non-LMM} satisfies
	$
	\A \geq \zero
	$, has two positive columns, and is incoherent (see Def.~\ref{def:incoh}). The dimensions satisfy $M\geq r \geq 3$.
	\item \label{as-S} The columns of $\S$ satisfy $\s_j \in \text{int}~\Delta_r, ~\forall j\in [N]$. Moreover, $\s_j$'s are sampled from a Dirichlet distribution with parameters $\boldsymbol{\mu} = [\mu_1, \mu_2, \cdots, \mu_r]$.
\end{enumerate}

For brevity, let us define a matrix function that has $\k$ acting on the columns of its matrix argument, $\boldsymbol{T_k}(\X) = [\k(\x_1), \k(\x_2), \cdots, \k(\x_N)]$ for $\X = [\x_1, \x_2, \cdots, \x_N]$. We are ready to state the following results.



\begin{Theorem}{(Main results)}\label{theorem}
	Under assumptions \ref{as-phi}, \ref{as-A}, \ref{as-S}, and supposing that 
	after performing a certain training procedure (see Section \ref{ch:learning}) on $f_1, f_2, \cdots, f_M$, the output satisfies 
	\begin{align}\label{eq:non-unm}
	\sum_{i=1}^{M}k_i(\a_i^\ttt\s)  = 1, ~ \forall \s \in \text{int}~\Delta_r.
	\end{align}
	Furthermore, assume that the composite functions $k_i$'s are all convex (or all concave).
	Then the following hold
	\begin{enumerate}[label=(\alph*)]
		\item The functions $k_1, k_2, \cdots, k_M$ are affine;
		\item The functions $\phi_1^{-1}, \cdots, \phi_M^{-1}$ are identified up to an affine transformation, i.e. $f_i(x) = d_i \phi_i^{-1}(x) + b_i, ~\forall i \in [M]$, where $d_i$'s and $b_i$'s are constants.
	\end{enumerate}
\end{Theorem}
The proof can be found in the appendix. A remark about function $\boldsymbol{T}_{\boldsymbol{k}}$ is in order.

\begin{Remark}
	According to (a) in Theorem~
	\ref{theorem}, we can write 
	\begin{align}\label{eq:tk_affine}
	\boldsymbol{T}_{\boldsymbol{k}}(\X) &= \D\X + \b\one_N^\ttt,
	\end{align}
	where $\D = \text{diag}(d_1, \cdots, d_M)$, and $\b = [b_1, \cdots, b_M]^\ttt$, and $d_i$ and $b_i$ are coefficients for the affine function $k_i$. Equation \eqref{eq:tk_affine} suggests that $\boldsymbol{T}_{\boldsymbol{k}}$ is an {affine} function in $\X$.
	However, we would like $\boldsymbol{T}_{\boldsymbol{k}}$ to be \emph{linear} in $\X$, instead of affine, as later we show that it is possible to identify parameters in LMM under invertible {linear} transformation  (Lemma~\ref{lemma:linear-LMM}).
	
	Fortunately, for signal model \eqref{eq:non-LMM} satisfying \ref{as-phi}, \ref{as-A} and \ref{as-S}, we can see that $\boldsymbol{T}_{\boldsymbol{k}}(\X)$ is indeed a linear function of $\X$. Let us consider a matrix $\X \in \R^{M\times N}$. 
	Due to equation \eqref{eq:non-unm}, we have $\one_M^\ttt\boldsymbol{T}_{\boldsymbol{k}}(\X) = \one_M^\ttt\D\X +  \one_M^\ttt\b\one_N^\ttt = \one_N^\ttt$, which means $\one_N^\ttt = \one_M^\ttt\D\X/(1-\one_M^\ttt\b)$. Plugging this into the above equation, we have
	\begin{align}
	\boldsymbol{T}_{\boldsymbol{k}}(\X) & = \D\X + \b \left( \frac{1}{1-\one^\ttt_M\b}\one_M^\ttt\D\X\right) \nonumber\\
	& = \left(\eye + \frac{1}{1-\one^\ttt_M\b}\b\one_M^\ttt\right)\D\X \nonumber\\
	&= \W\X
	\end{align}
	where we define $\W := \left(\eye + \frac{1}{1-\one^\ttt_M\b}\b\one_M^\ttt\right)\D$, and $\one_M$ is an all-one vector of length $M$.  The above equation suggests that $\boldsymbol{T_k}$ is \emph{linear} in $\X$. A subtle point is that the above calculation is invalid when $1=\one^\ttt_M\b$ holds \emph{exactly}, but this is extremely unlikely since $\b$ will be resulted from a numerical algorithm. 	
\end{Remark}

We will propose a method to make \eqref{eq:non-unm} (approximately) hold in Section~\ref{ch:learning}. Let us briefly discuss the roles of the assumptions.
For \ref{as-phi}, the invertibility condition is important, as one in general cannot hope to recover the unknown parameters if they undergo non-invertible transformations. The twice differentiable condition on $\phi_i$'s is to make $k_i$'s twice differentiable, when suitable $f_i$'s are learned. This is also natural, as it requires the nonlinear functions in data generation to be smooth. 

Assumption \ref{as-A} is the same as in Lemma~\ref{lemma-dr}, except for the additional incoherent assumption. 
The incoherence assumption is important, as it ensures that solutions that satisfy \eqref{eq:non-unm} exist, see detailed discussion in Section~\ref{ch:feasibility}. The condition that it should have two positive columns may seem strange, but it is easily satisfied if, say, $\A$ is generated from an absolutely continuous distribution, supported on the nonnegative orthant. For \ref{as-S}, the Dirichlet distribution is assumed because it gives samples on the probability simplex. In addition, this assumption ensures that the columns of $\S$ cover the entire interior of $\Delta_r$ as $N\rightarrow +\infty$, which plays a role when characterizing the asymptotic {identification guarantee} of the proposed method as in Corollary~\ref{theorem:inf}. 

Given the generative model \eqref{eq:non-LMM}, Theorem~\ref{theorem} essentially asserts that if we require $\one^\ttt\y =1$ for all input $\s$, then the learned functions $f_1, \cdots, f_M$ will remove the nonlinearity in functions $\phi_1, \cdots, \phi_M$. But our main goal is identifying parameters in the latent LMM; $\boldsymbol{T}_{\k}$ being linear is not enough. To see this more clearly, suppose we get a solution for $f_i$'s of this form
\begin{align}\label{eq:trivial}
f_i(x) = 1/M, ~~i\in [M].
\end{align}
In this case, $k_i$'s are all constant functions, and hence convex. Moreover, for this solution \eqref{eq:trivial}, we have $\k(\A\s) = \D\A\s+ \boldsymbol{b}$, where $\D = \zero$ and $\b = (1/M) \one$; meaning that $\boldsymbol{f}$ maps all input $\x =\boldsymbol{\phi}(\A\s)$ to the single point $\y=(1/M)\one$, which does satisfy \eqref{eq:non-unm}.

The problem we identify here is important: we need additional constraints on $\y$ beyond $\one^\ttt\y =1$, so that $\y$ preserves information about the original data $\x$, as only then we can hope to identify $\A$ and $\s$ from $\y$. We propose a method to remedy this in Section~\ref{ch:learning}. 

To proceed with parameter estimation, let us provide the following lemma, concerning parameter identifiability of LMM \eqref{eq:LMM} under a linear transformation.
\begin{Lemma}\label{lemma:linear-LMM}
	Consider the LMM model $\X=  \A\S$, where $\A \in\R^{M\times r}$ and $\S\in\R^{r\times N}$ satisfies the SS condition, and $\text{rank}(\A) = \text{rank}(\S) = r$. Let $\Y = \W\X$, where $\W \in \R^{M\times M}$ is nonsingular. Then we can identify $\widetilde{\A} = \W\A$ and $\S$ up to column permutation by solving
	\begin{align}\label{eq:lin_volmin}
	\min_{\B \in \R^{M\times r}, \H \in \R^{r\times N}} &~~ \text{Vol}(\B) \nonumber\\
	\st &~~ \Y = \B\H, \nonumber\\
	&~~ \H \geq {\bf 0}, ~ \H^\ttt \one  = \one.
	\end{align}
	That is, suppose $(\B^*, \H^*)$ is an optimal solution of the above problem, then $\B^* = \widetilde{\A}\boldsymbol{\Pi}$ and $\H^* = \boldsymbol{\Pi}^\ttt \S$, where $\boldsymbol{\Pi}$ is a permutation matrix.
\end{Lemma}
This lemma is a direct consequence of Theorem~\ref{thm:volmin-ident}. It suggests when the original model $\X = \A\S$ is identifiable, then  after an invertible linear transformation $\W$, we can still identify $\S$ using VolMin; but it is not possible to identify $\A$ due to the linear transformation $\W$. This lemma also suggests that we can employ an algorithm designed to tackle LMM to identify $\S$, once the nonlinear effects in \eqref{eq:non-LMM} have been removed, and only an unknown linear transformation is left.

\subsection{Feasibility of \eqref{eq:non-unm}}\label{ch:feasibility}
Results in Theorem~\ref{theorem} hinge on equation \eqref{eq:non-unm}. One could be wondering, giving the conditions outlined in assumptions \ref{as-phi}, \ref{as-A}, and \ref{as-S}, does there exist $\boldsymbol{f}$ such that \eqref{eq:non-unm} hold? This amounts to 
study feasibility of \eqref{eq:non-unm}, which is not obvious. For instance, consider the \emph{naturally} guessed solution $\{\widehat{f}_i = \phi^{-1}_i, ~\forall i \}$, for which we have $\boldsymbol{T}_{\boldsymbol{k}}(\X) = \X$; but we \emph{don't} have $\sum_{i=1}^{M}k_i(\a_i^\ttt\s) = \sum_{i=1}^{M}\a_i^\ttt\s = 1, ~\forall s\in \text{int}~\Delta_r$ without imposing more restrictive assumptions on $\A$ or $\S$. This means that, for this natural guess, \eqref{eq:non-unm} does not hold.

To study this feasibility issue, we note that if there exists a diagonal matrix $\D$, such that $\one^\ttt\D\A = \one^\ttt$, then letting $\widetilde{f}_i = \phi_i^{-1}$, we have
\begin{align}
\sum_{i=1}^{M}d_i\widetilde{f}_i(\phi_i(\a_i^\ttt\s)) &= \sum_{i=1}^{M} d_i\a_i^\ttt\s \nonumber\\
&= \one^\ttt\D\A\s \nonumber\\
&= \one^\ttt\s \nonumber\\
&= 1 \qquad \forall \s\in \text{int}~\Delta,
\end{align}
where $d_i$ is the $i$-th diagonal element of $\D$. Hence, the functions $\left\lbrace \widehat{f}_i(\cdot) = d_i\widetilde{f}_i(\cdot), ~i\in[M]\right\rbrace $ satisfy \eqref{eq:non-unm}. An additional requirement is that $\{d_i \neq 0, \forall i\}$, otherwise we can get a trivial solution, as explained in the above section.

Building on the above observation, the feasibility problem of \eqref{eq:non-unm} boils down to establishing existence of a \emph{nonsingular} diagonal matrix $\D$ (i.e. $d_i\neq 0, \forall i$), such that $\one^\ttt\D\A = \one^\ttt$, for matrix $\A$ that satisfies assumption \ref{as-A}. 
We present Proposition~\ref{prop:sto}, which shows that with a mild incoherence condition (see Definition~\ref{def:incoh}) on $\A$, such desired $\D$ indeed exists. We start by providing the following definition of incoherence.

\begin{Def}{(Incoherence)}\label{def:incoh}
	A tall and full-rank matrix $\A\in \R^{m\times r}$ is a said to be incoherent if $\boldsymbol{e}_j \notin \text{Range}(\A), ~\forall j\in [m]$.
\end{Def}
Note that here incoherence is defined in the same spirit as the incoherence found in well-known compressed sensing literature, see e.g. \cite{candes2009exact}.

We are now ready to state the following proposition. Here we write $\A^\ttt\d = \one_r$ instead of $\one^\ttt\D\A = \one^\ttt$ for conciseness: existence of nonsingular diagonal $\D$ is the same as existence of fully dense $\d$. 
\begin{Prop}\label{prop:sto}
	For a tall, full rank, and incoherent matrix $\A\in \R^{m\times r}$, there exists a vector $\d\in\R^m$, such that
	\begin{subequations}
		\begin{align}
		\A^\ttt\d &= \one_r,\label{eq:sto}\\
		\|\d\|_0 &= m.\label{eq:fully-dense}
		\end{align}	
	\end{subequations}
\end{Prop}
Note that by assumption, $\A$ is tall and full rank, so there are infinitely many $\d$ vectors satisfy \eqref{eq:sto}. However, it is not obvious if there is always a \emph{fully dense} $\d$ (i.e. \eqref{eq:fully-dense}) such that \eqref{eq:sto} holds for any $\A$ that is tall and full rank.

The proof of Proposition \ref{prop:sto} can be found in appendix.



\begin{Remark}
	We establish that for an incoherent $\A$, there always exist solutions to make \eqref{eq:non-unm} hold. Moreover, we point out that even for some $\A$ that is not incoherent, solutions for \eqref{eq:non-unm} might also exist. For example, if one or more columns of $\A$ are some columns of an identity matrix, then $\A$ is not incoherent. However, if we have $\one^{\ttt}\A=\one^{\ttt}$ -- which is true when all columns of $\A$ are some columns of an identity matrix -- then we see that $\{f_i = \phi_i^{-1}, ~\forall i\}$ is a feasible solution.
\end{Remark}

\subsection{Learning algorithm}\label{ch:learning}

Theorem \ref{theorem} suggests the following optimization formulation to learn desired $\boldsymbol{f}$
\begin{align}\label{eq:regress-find-inf}
\text{find} &~~ f_1, \cdots, f_M \nonumber \\
\st &~~  f_i\circ \phi_i ~\text{is all convex (or all concave)} ~\forall i \in [M], \nonumber \\ &~~\sum_{i=1}^{M}f_i(\x_j(i)) =1 ~\forall j \in [N].
\end{align}
For this formulation we have the following claim.
\begin{Corollary}\label{theorem:inf}
	For problem \eqref{eq:regress-find-inf}, suppose the data $\X = [\x_1, \cdots, \x_N] \in \R^{M\times N}$ admit model \eqref{eq:non-LMM} and assumptions \ref{as-phi}, \ref{as-A}, \ref{as-S} hold.
	Suppose $N\rightarrow +\infty$, the optimal solutions to \eqref{eq:regress-find-inf} satisfy \eqref{eq:non-unm}, and the resulting $\{k_i = f_i \circ \phi_i,~\forall i\in [M]\}$ are all affine.
\end{Corollary}
This corollary follows from the distributional assumption \ref{as-S} on $\s_j$. As $N\rightarrow +\infty$, $\s_j$ will cover all the { interior} of $\Delta_r$ with probability 1. Then the constraints in \eqref{eq:regress-find-inf} become the same as the conditions in Theorem~\ref{theorem}. Corollary~\ref{theorem:inf} thus guarantees the {nonlinear function identification property} of formulation \ref{eq:regress-find-inf} in an \emph{asymptotic} sense. In the following, we approximate problem \ref{eq:regress-find-inf} to make it amenable to numerical algorithms. In 
Section~\ref{ch:experiments}, we give numerical examples, showing that even with \emph{finite} $N$, the proposed method works remarkably well.

Problem formulation \ref{eq:regress-find-inf} suggests that we need to find functions $f_1, \cdots, f_M$, such that the output sums to one. To enforce the constraint that $k_i$'s are all convex (or all concave), we note
\begin{align}
k_i''(x) = f_i''(\phi_i(x))[\phi_i'(x)]^2 + f_i'(\phi_i(x))\phi_i''(x).
\end{align}
To make sure $k_i$ is convex (or concave), we need $k_i''(x) \geq 0$ (or $k_i''(x) \leq 0$), which requires us to know the sign of $\phi_i''(x)$. For instance, suppose $\phi_i''(x) \leq 0$, then we can pick a parametric family for $f_i$'s, such that $f_i''(x) \geq 0$ and $f_i'(x) \leq 0$. Then we have $k_i''(x) \leq 0$, i.e. $k_i$ is concave. Similarly, we can constrain $f_i$'s for all $i\in [N]$ to make sure $k_i$'s are all convex (or concave). To simplify implementation, we adopt an approximation: We only require $f_i$'s to be invertible in this work.
This leads to the following optimization problem.
\begin{align}\label{eq:regress-find}
\text{find} &~~ f_1, \cdots, f_M \nonumber \\
\st &~~  f_i ~\text{is invertible} ~\forall i \in [M], \nonumber \\ &~~\sum_{i=1}^{M}f_i(\x_j(i)) =1 ~\forall j \in [N].
\end{align}
In other words, we aim at learning \emph{invertible} functions that add to one. The invertibility condition is crucial, otherwise we can obtain trivial solutions, as explained before. 


To parametrize functions $f_j$, we will adopt Neural Networks (NN) with one hidden layer, due to their universal approximation capability \cite{hornik1989multilayer, barron1993universal}.
In particular, we employ the following parametric function family
\begin{align}\label{eq:nn-funcs}
\begin{split}
\mathcal{F} = \left\lbrace f \Bigg \lvert f(x) = \sum_{k=1}^{K}\alpha_k \sigma(\beta_kx + \gamma_k) + \delta_k, \right. \\ 
\left. ~~\alpha_k>0, ~\beta_k >0, ~\forall k  \in [K] \vphantom{\Bigg \lvert} \right\rbrace
\end{split}
\end{align}
where $K$ is the number of neurons, $\{\alpha_k, \beta_k, \gamma_k, \delta_k\}_{k=1}^K$ are the learnable parameters of this NN, and $\sigma$ denotes the nonlinearity. Importantly, the constraints on $\alpha_k$ and $\beta_k$ are to ensure invertibility, as stated below.
\begin{Lemma}
	In \eqref{eq:nn-funcs}, if $\sigma'(x) > 0, ~\forall x$,  the functions in $\mathcal{F}$ are all invertible.
\end{Lemma}
The above lemma can be easily seen to be true. By definition, we have $f'(x) = \sum_{k=1}^{K}\alpha_k\beta_k \sigma'(\beta_kx+ \gamma_k)$. 
For $\sigma'(x) >0$, we have $f'(x) > 0$ if $\alpha_k>, ~ \beta_k > 0, ~\forall k \in [K]$.
Note that the requirement for $\sigma'(x) > 0$ is easily satisfied for commonly used neurons, e.g., $\text{tanh}(\cdot)$ and the sigmoid function. For this reason, we pick $\sigma$ as $\text{tanh}(\cdot)$ in this work.
\begin{figure*}[th]
	\centering
	\includegraphics[width=.95\textwidth]{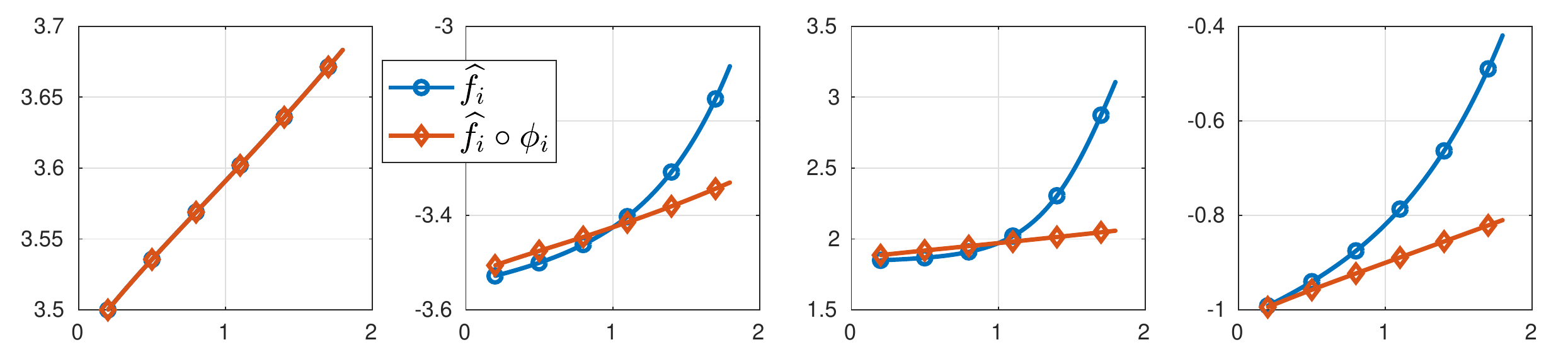}
	\vspace{-10pt}
	\caption{Learned functions and their composition with the ground truth nonlinear functions used for data generation. The four functions for data generation are $\phi_1(x) = x$, $\phi_2(x) = \sqrt{x}$, $\phi_3(x) = \sqrt[4]{x}$, $\phi_4(x) = \log(x+1)$.  The $\phi_i$'s are kept secret in the learning stage.}
	\label{fig:visual}
\end{figure*}

Utilizing the parametric family $\mathcal{F}$ in \eqref{eq:nn-funcs}, we arrive at the following optimization problem
\begin{align}\label{eq:opt}
\min_{\{ \alpha_k^i, \beta_k^i, \atop \gamma_k^i, \delta_k^i\} }&~~ \frac{1}{N}\sum_{j=1}^{N}\left(1 - \sum_{i=1}^{M} \sum_{k=1}^{K}\alpha_k^i \sigma(\beta_k^i\x_j(i) + \gamma_k^i) + \delta_k^i\right)^2 \nonumber \\
\st &~~ \alpha_k^i>0, ~\beta_k^i>0,\quad \forall k \in [K], ~i \in [M].
\end{align}
This is a nonlinear least-squares regression problem, with bound constraints. We employ a trust-region algorithm \cite{coleman1996interior} for optimization.

After obtaining parameters $\{\widehat{\alpha}_k^i, \widehat{\beta}_k^i, \widehat{\gamma}_k^i, \widehat{\delta}_k^i\} $ via \eqref{eq:opt}, we obtain $\widehat{f}_i(x) = \sum_{k=1}^{K} \widehat{\alpha}_k^i \sigma(\widehat{\beta}_k^i x + \widehat{\gamma}_k^i) + \widehat{\delta}_k^i$, and form the transformed data $\Y = \boldsymbol{\widehat{f}}(\X)$. Theorem \ref{theorem} predicts that $\Y \approx \W\A\S$ for some nonsingular matrix $\W$. From Lemma \ref{lemma:linear-LMM}, we see that we can employ an algorithm for LMM to identify $\S$. For this purpose, we employ the classical MVES algorithm \cite{chan2009convex} for LMM, and obtain an estimate~$\widehat{\S}$.

The overall procedure is summarized in Algorithm \ref{alg}. We {emphasize} again that the method is \emph{unsupervised}: The only data is $\X$, not $\{\x_j, y_j \}_{j=1}^N$ (feature-label pairs) as in, e.g., the generalized additive models \cite[Ch.~9]{Hastie2009elements} setting, or recent works on nonlinear estimation \cite{yi2015optimal, chen2017sparse}.
\begin{algorithm}
	\begin{algorithmic}[1]
		\REQUIRE Data $\X \in \R^{M\times N}$, number of neurons $K$, latent dimension $r$
		\ENSURE Learned functions $\widehat{f}_1, \cdots, \widehat{f}_M$, estimated $\widehat{\S}$
		\STATE Learn parameters $\{\widehat{\alpha}_k^i, \widehat{\beta}_k^i, \widehat{\gamma}_k^i, \widehat{\delta}_k^i\} $ by solving \eqref{eq:opt}
		\STATE Form functions $\widehat{f}_1, \cdots, \widehat{f}_M$ by $\widehat{f}_i(x) = \sum_{k=1}^{K} \widehat{\alpha}_k^i \sigma(\widehat{\beta}_k^i x + \widehat{\gamma}_k^i) + \widehat{\delta}_k^i$ 
		\STATE Obtain transformed data by applying the learned functions on input data: $\Y = \boldsymbol{\widehat{f}}(\X)$
		\STATE Obtain $\widehat{\S}$ by calling $\text{MVES}(\Y, r)$
		\RETURN $\widehat{f}_1, \cdots, \widehat{f}_M$, $\widehat{\S}$
	\end{algorithmic}
	\caption{Nonlinear matrix factor recovery}
	\label{alg}
\end{algorithm}

\section{Numerical experiments}\label{ch:experiments}

\subsection{Synthetic data study}
{We start by providing} a qualitative assessment of the proposed theory and algorithm. For this purpose, we will visualize the learned functions to see if nonlinearity in data generation is indeed resolved. We randomly generate $\S$ according to a Dirichlet distribution -- such that the generated $\s_j$'s are nonnegative and sum to one. The dimensions are $M = r = 4$ and $N=1000$. The parameter of this Dirichlet distribution is set to $\boldsymbol{\mu} = [0.1, 0.1,0.1,0.1]$, so that the generated $\s_j$'s are well spread on the probability simplex, {hence SS is likely to be satisfied}.
For this experiment, we take $\A$ to be $\A = 2\boldsymbol{I}_4$. 
The four nonlinear functions in data generation are $\phi_1(x) = x$, $\phi_2(x) = \sqrt{x}$, $\phi_3(x) =  \sqrt[4]{x}$, and $\phi_4(x) = \log(x+1)$. Note that these functions are \emph{not} revealed to the learning algorithm, and are only used to visualize the results after learning is completed. For learning, each function $f_i$ is parametrized by a constrained one-hidden-layer NN defined in  \eqref{eq:nn-funcs}, with $K=20$ neurons. The learned functions $f_1 \cdots f_4$ and the composite functions $f_1\circ \phi_1 \cdots f_4 \circ \phi_4$ are shown in Figure~\ref{fig:visual}. 

One can immediately see that the learned functions indeed resolve nonlinearity in data generating nonlinear functions: The learned $f_1$ is a linear function since $\phi_1$ is a linear function; the other learned functions all look similar to the corresponding inverse functions of $\phi_i$'s. Moreover, one can clearly see that the composite functions all look affine.

\begin{figure*}[ht]
	\centering
	\includegraphics[width=.95\textwidth]{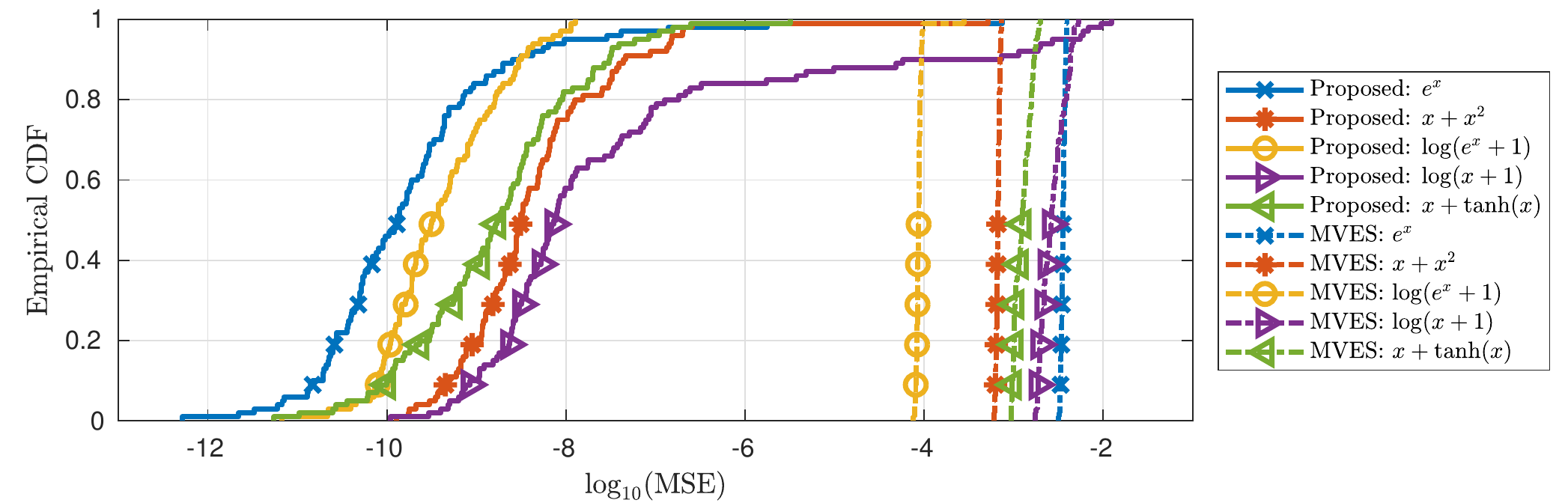}
	\vspace{-5pt}
	\caption{Empirical CDF of MSE: the legend shows the learning method, and the nonlinear function used in data generation. For each nonlinear function, $100$ trials are generated. A point $(-6,0.99)$ on a curve means the corresponding learning method yields $\text{MSE} \leq 10^{-6}$ in $99\%$ of the $100$  trials.}	
	\label{fig:cdf}
	\vspace{-5pt}
\end{figure*}

Next, we test the parameter estimation performance. For this experiment, we generate data with five different nonlinear functions: 
\begin{enumerate*}[label={(\alph*)}]
	\item $e^x$,
	\item $x+x^2$,
	\item $\log(e^x+1)$,
	\item $\log(x+1)$,
	\item $x+\text{tanh}(x)$.
\end{enumerate*}
For each case, one of the five functions are used for \emph{all} coordinates (features), i.e. $\phi_1 = \cdots=\phi_M$.
The parameter settings are $M=10$, $N=1000$, and $r=4$. 
We generate $\A\in \R^{10\times 4}$ by sampling a standard normal distribution for each entry, and then take the absolute values, followed by a column normalization step.
$\S$ is similarly generated as in the first experiment. For this experiment,  the $f_i$ functions are constrained to be the same: a constrained one-hidden-layer NN defined in \eqref{eq:nn-funcs}, with $K=40$ for all cases, to avoid unrealistic parameter tuning. In other words, all the NN share the same parameters. Since problem \eqref{eq:opt} is nonconvex, different initialization could lead to different results. For this reason, the formulation \eqref{eq:opt} is optimized five times with different random initialization, and the result of \emph{smallest cost function value} is used for subsequent steps of Algorithm 1. 
The performance metric we employ is mean squared error (MSE): $\text{MSE} = \frac{\|\widehat{\S} - \S \|_F^2}{rN} $.

Since our method is the first work dealing with this nonlinear model, 
{the only baseline we employ is MVES without considering nonlinear effects.}
The motivation is to see if it is indeed possible to estimate parameters with \emph{unknown} nonlinear functions, {using} only nonlinearly distorted data $\X$.
For each setting, $100$ trials with different randomly generated data (see appendix for details) are performed, and the empirical cumulative distribution function (CDF) of the resulting MSEs are reported in Figure~\ref{fig:cdf}. 

From Figure~\ref{fig:cdf}, one can see that the proposed method yields {significant} improvements over applying MVES directly, in all the cases. Note that the x-axis in Figure~\ref{fig:cdf} is $\log_{10}(\text{MSE})$, hence our method yields several order of magnitude improvement in accuracy over the baseline. There are a few trials where the proposed method yields relatively larger error, which is likely caused by numerical difficulties in optimizing NNs.

\begin{figure}
\begin{subfigure}[b]{0.49\linewidth}
	\begin{overpic}[width=\linewidth]{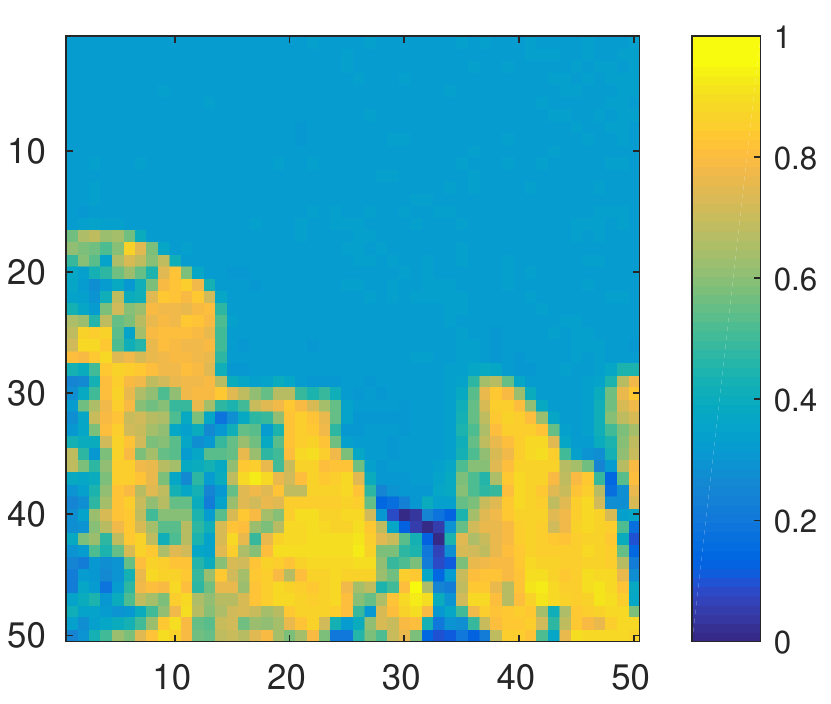}
		\put(40, 50){water}
		\put(36, 16){soil}
	\end{overpic}
\end{subfigure}
\begin{subfigure}[b]{0.49\linewidth}
	\begin{overpic}[width=\linewidth]{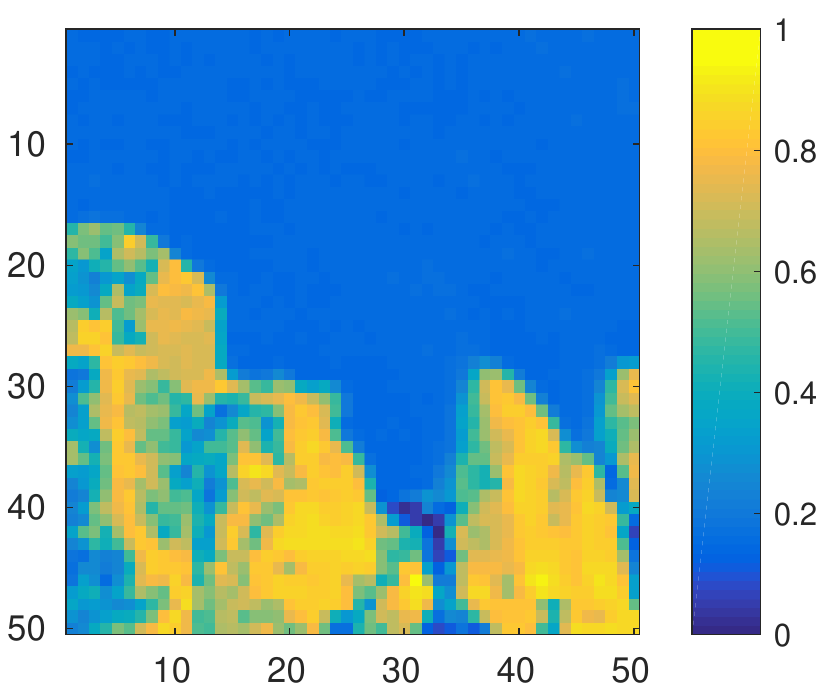}
		\put(40, 50){water}
		\put(36, 16){soil}
	\end{overpic}
\end{subfigure}	
	\caption{Estimated $\widehat{\bm S}(2,:)$ (soil map) by MVES (left) and the proposed method (right). Text in the figure indicates ground-truth.}
	\label{fig:hu}
	\vspace{-0.5cm}
\end{figure}

\subsection{Case study with a hyperspectral image}
We next perform an experiment on hyperspectral unmixing (HU). Unlike normal RGB images, a pixel in a hyperspectral image contains information on hundreds of spectral bands. With the more detailed spectral information, it is reasonable to assume that different materials have their distinct spectral signature. Physically, each pixel represents a convex combination of materials that are present for the geographical region. However, it is known that the collected measurement may encounter nonlinear distortion.
The HU task involves separating materials of a ground region.

The image employed in this experiment is the Moffett Field captured in France -- a standard benchmark for testing HU algorithms. The region has three main materials: water, soil, and vegetation. This scene is known for the existence of nonlinear mixture pixels - which usually poses a challenge to LMM-based HU algorithms such as MVES. The size of the image is $50\times 50$, hence we have 2500 pixels. Each pixel is measured on 224 spectral bands. Following commonly applied preprocessing steps \cite{fu2016robust}, we remove the water-absorbing bands, and end up with a matrix $\X$ of size $200 \times 2500$, so that each of the remaining 200 spectral bands serves as a feature for that pixel. The algorithms are supposed to identify what materials are present in each pixel, and the proportion of the presenting materials.

To apply our method, we use the same $f_i$ on each of the $200$ feature as above, and fix $K=40$.
We compare our method with MVES, since MVES is one of the best performing methods for HU. After obtaining the estimated $\S$, we inspect each row of $\S$ to determine which of them corresponds to the water, soil, and vegetation portion of the image.
The difference between the two sets of results is most visible in the estimated soil distribution (a particular row of estimated $\S$) as shown in Figure~\ref{fig:hu}: the result by MVES outputs large values in the water region.
The proposed method outputs much smaller values in the water region, which is much more aligned with reality. 

\begin{figure}[ht]
	\centering
	\begin{overpic}[width=.8\linewidth]{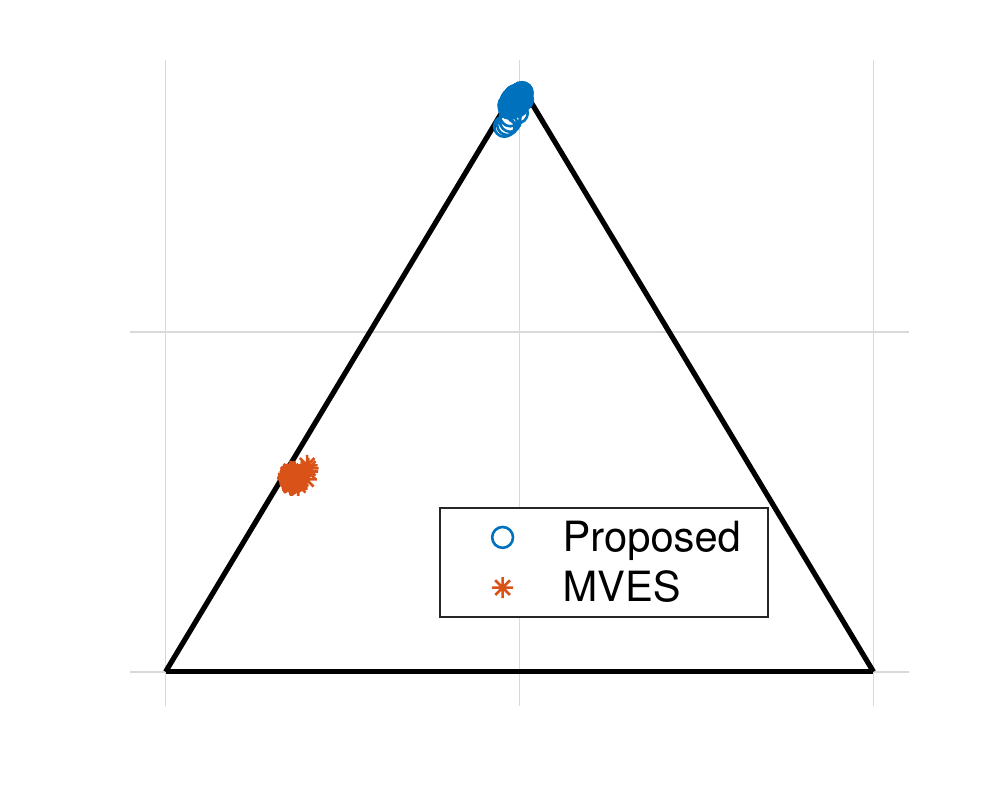}
		\put(5, 3){$[1, 0, 0]^\ttt$}
		\put(55, 70){$[0, 0, 1]^\ttt$}
		\put(80, 3){$[0, 1, 0]^\ttt$}
	\end{overpic}
	\caption{Visualizing columns of the estimated $\S$ corresponding to the water region.}
	\label{fig:S_est}
\end{figure}

We further plot the estimated $\S$ in the known water region (top $15 \times 50$ part\footnote{We take this part as it is clear that there is only one material (water) in this region, so the ground truth for each column of $\S$ is any permutation of $[1, 0, 0]^\ttt$.} of Figure~\ref{fig:hu}), as shown in Figure~\ref{fig:S_est}. Since columns of $\S$ live in a dimension-2 simplex, we project all the points into a 2D space, with the vetices of the triangle corresponding to the original vetices in the 3D space, as shown in Figure~\ref{fig:S_est}.
Note that Figure~\ref{fig:hu} shows a \emph{single} estimated row of $\widehat{\S}$ for easy visualization, while Figure~\ref{fig:S_est} presents results from all rows, for the part that corresponds to the top $15\times 50$ region. 
From this figure, we see that results of the proposed method coalesce around a coordinate vector $[0,0,1]^\ttt$, which means that
proposed method is quite certain that there is only one material in this region (which is true); while MVES is much less confident, as the points are much far away from a coordinate vector.
The estimated $\S$ also indicates that MVES fails to clearly separate soil and water spectral signatures (columns of $\A$), whereas our method performs much better.

\section{Conclusion}

This work serves as a first attempt to unravel latent structures in data when the observations are distorted with \emph{unknown} nonlinear effects. It is an important problem to consider in practice, but a concrete study is solely missing prior to this work. Much to one's surprise, this seemingly impossible mission of figuring out unknown nonlinearities can actually be accomplished up to affine transformations, as we showed in this paper. A learning algorithm based on the powerful artificial neural networks is proposed to rectify the unknown nonlinear functions. Our carefully designed numerical experiments show clear advantage in terms of inverting nonlinear distortions and identifying latent factors in LMMs altered by unknown nonlinear effects.

\clearpage
\bibliographystyle{plain}
\bibliography{ref.bib}

\clearpage
\section*{Appendix: ``Learning Nonlinear Mixtures: Identifiability and Algorithm''}

\section*{Some definitions in convex geometry}
\begin{Def}{(Convex cone)}
	The convex cone of $\{\x_1, \cdots, \x_N \}$ is defined as 
	\begin{align}
	&\text{cone}\{\x_1, \cdots, \x_N \} = \nonumber \\
	&\left\lbrace \x \Bigg\lvert \x = \sum_{j=1}^{N}\x_j\theta_j, ~\theta_j\geq 0, \forall j\in [N] \right\rbrace.
	\end{align}
\end{Def}

\begin{Def}{(Convex hull)}
	The convex hull of $\{\x_1, \cdots, \x_N \}$ is defined as 
	\begin{align}
	&\text{conv}\{\x_1, \cdots, \x_N \} = \nonumber \\
	& \left\lbrace \x \Bigg\lvert \x = \sum_{j=1}^{N}\x_j\theta_j, ~\sum_{j=1}^{N}\theta_j=1, ~\theta_j\geq 0, \forall j \in [N]\right\rbrace.
	\end{align}
\end{Def}

\begin{Def}{(Simplex)}
	A convex hull $\text{conv}\{\x_1, \cdots, \x_N \}$ is called a simplex if $\x_1, \cdots, \x_N$ are affinely independent, i.e., $\x_1 -\x_N, \cdots, \x_{N-1} - \x_N$ are linearly independent.
\end{Def}

A probability simplex is a special simplex, with all vertex vectors being the {coordinate} vectors, 
i.e. $\forall i\in [N], ~\x_i = \boldsymbol{e}_j$ for some $j$, where $\boldsymbol{e}_j$ has $1$ at its $j$-th coordinate, and $0$ for all other coordinates. 

\section*{Proofs}
\textbf{\emph{Proof of Lemma~\ref{lemma-dr}}}: Assume without loss of generality that the two nonzero columns are the first and second column. Let us denote 
\begin{align}\label{eq:lemma2}
\zeta(s_1, s_2,\cdots, s_{r-1}) := \sum_{i=1}^{M} \psi_i\left(\a_i^\ttt\s \right) =1, ~ \s \in \text{int}~\Delta_r.
\end{align}
Note that $\zeta$ is a function of $(r-1)$ variables $s_1, \cdots, s_{r-1}$, since $\one^\ttt\s = 1$.
Equation \eqref{eq:lemma2} suggests that $\zeta$ is a constant function on $\Delta_r$. Taking derivative with respect to (w.r.t.) $s_1$ and $s_2$, we get
\begin{align}
\frac{\partial \zeta}{\partial s_1} = \sum_{i=1}^{M}\a_i(1)  \psi_i'\left( \a_i^\ttt\s\right),
\end{align}
and 
\begin{align}\label{eq:dr_3}
\frac{\partial^2 \zeta}{\partial s_1 \partial s_2} &= \sum_{i=1}^{M}\a_i(1) \a_i(2) \psi_i''\left(\a_i^\ttt\s \right)= 0.
\end{align}
By the assumption on $\A$, we have $\a_i(1)\a_i(2) > 0, ~\forall i$. The assumption that $\psi_i$'s are all convex (or concave) translates to $\psi_i'' \geq 0$ (or $\psi_i'' \leq 0$), for all $i\in [M]$. From \eqref{eq:dr_3}, we conclude that $\psi_i''=0,~\forall i$, which suggests that all the $\psi_i$'s are affine.  $\blacksquare$


While we prove the above lemma for our use in this work, more results concerning functional equations can be found in several books on this topic, see e.g. \cite{kuczma2009introduction, efthimiou2010introduction}.

\textbf{\emph{Proof of Theorem~\ref{theorem}:}}
Given assumptions (A2) and equation \eqref{eq:non-unm}, (a) is a direct consequence of Lemma 2. 

For (b), we note that from (a), $k_i(t) = d_it + b_i$ for some constants $d_i$ and $b_i$. Let $x = \phi_i(t)$, then $t = \phi^{-1}(x)$.
Plugging into $f_i(\phi_i(t)) = d_it+b_i$, we obtain $f_i(x) = d_i\phi^{-1}(x) + b_i$.

To prove Proposition \ref{prop:sto}, we need Lemma~\ref{lemma:incoh} and Lemma~\ref{lemma}, which are presented here and their proof will follow.
\begin{Lemma}\label{lemma:incoh}
	Suppose $\A\in \R^{m\times r}$ is full rank and incoherent, i.e. $\boldsymbol{e}_i \notin \text{Range}(\A), \forall~ i\in [m]$. Then $\widehat{\A} =\left[\begin{array}{c}
	\A \\
	\one_r^\ttt
	\end{array}\right] $ is incoherent.
\end{Lemma}
This lemma asserts that if a matrix $\A$ is incoherent, then appending a row of all 1's preserves incoherence.

\begin{Lemma}\label{lemma}
	For a tall and full rank matrix $\A\in \R^{m\times r}$, where $\A$ is incoherent, there exists a $\d\in\R^m$, such that 
	\begin{subequations}
		\begin{align}
		\A^\ttt\d &= \zero_r, \label{eq:stz} \\
		\|\d\|_0 &= m.
		\end{align}
	\end{subequations}
\end{Lemma}

\textbf{\emph{Proof of Lemma~\ref{lemma:incoh}:}}
The incoherence condition means that there is no such $\y \in \R^r$, such that $\A\y = \boldsymbol{e}_i$ for any $i \in [m]$.
Suppose there is a $\widehat{\y}\in \R^r$, such that $\widehat{\A}\widehat{\y} = \boldsymbol{e}_j$ for some $j\in [m+1]$. There are two cases
\begin{enumerate}
	\item $1 \leq j\leq m$:  This means we have $\widehat{\y}$ such that $\A\widehat{\y} = \boldsymbol{e}_j$ for some $j\in [m]$ -- a contradiction to the assumption that $\A$ is incoherent.
	\item $j = m+1$: This means that $\A\widehat{\y} = \zero_m$ for $\widehat{\y}\neq \zero_r$-- a contradiction to the assumption that $\A$ is full rank.
\end{enumerate}
Hence  $\widehat{\A}$ is incoherent if $\A$ is full rank and incoherent. $\blacksquare$

\textbf{\emph{Proof of Lemma~\ref{lemma}:}}
Let $\U \in \R^{m\times (m-r)}$ be a set of bases of the null space of $\A$, i.e. 
\begin{align}
\text{Range}(\U) = \text{Null}(\A).
\end{align}
By assumption, $\A$ is incoherent, hence $\boldsymbol{e}_j \notin \text{Range}(\A), ~\forall j\in [m]$. For any $j$, we have the decomposition
\begin{align}
\boldsymbol{e}_j = \widehat{\boldsymbol{e}}_j + \overline{\boldsymbol{e}}_j,
\end{align}
where $\widehat{\boldsymbol{e}}_j \in \text{Range}(\A)$ and $\overline{\boldsymbol{e}}_j \in \text{Range}(\U)$. Since $\boldsymbol{e}_j \notin \text{Range}(\A)$, we have $\boldsymbol{e}_j^\ttt\U = \overline{\boldsymbol{e}}_j^\ttt\U \neq \zero_{m-r}, ~\forall j\in [m]$, which means $\U$ does not have a row that is all-zero.

Let $
\mathcal{I}_1, \cdots, \mathcal{I}_{m-r}$ be the index sets of nonzero entries in each column of $\U$, then we have $\cup_{j=1}^{m-r}\mathcal{I}_j = [m]$ since $\U$ does not have an all-zero row. Let us present the following useful fact.
\begin{Fact}\label{fact}
	Let $\x, \y \in \R^m$, with sets $\mathcal{I}_{\x}$ and $\mathcal{I}_{\y}$ being the sets of indices of nonzero entries, then we can find a vector $\boldsymbol{z} \in \text{Span}\{\x, \y \}$, such that $\mathcal{I}_{\z} = \mathcal{I}_{\x} \cup \mathcal{I}_{\y}$.
\end{Fact}
\emph{Proof:} Let $a = \frac{1}{\max_j |\x_j|}$ and $b = \frac{2}{\min_{j: \y_j \neq 0} |\y_j|}$. The denominator of $b$ is the minimum of absolute value of the nonzero entries of $\y$. Consider the vector 
\begin{align}
\z = a\x + b\y.
\end{align}
By the choice of $a$ and $b$, we have $\max_j |a\x_j| =1$ and $\min_{j: \y_j\neq 0} |b\y_j| = 2$. Hence for any $j$ where $\x_j \neq 0$ and $\y_j\neq 0$, we have $a\x_j + b\y_j\neq 0$. This shows that there exists a $\z \in \text{Span}\{\x, \y \}$, such that $\mathcal{I}_{\z} = \mathcal{I}_{\x}\cup \mathcal{I}_{\y}$. $\blacksquare$

We can now utilize Fact \ref{fact} to show that there exists a fully dense $\d \in \text{Range}(\U)$. Consider the first two columns of $\U$: $\U_1$ and $\U_2$. From Fact \ref{fact}, we can find a vector ${\u}\in \text{Span}\{ \U_1, \U_2 \}$, such that $\mathcal{I}_{\u} = \mathcal{I}_1 \cup \mathcal{I}_2$. Now consider $\u$ and $\U_3$, invoking Fact~\ref{fact} again, we can find a vector $\overline{\u} \in \text{Span}\{\u, \U_3 \}$, such that $\mathcal{I}_{\overline{\u}} = \mathcal{I}_{\u} \cup \mathcal{I}_3 = \mathcal{I}_1\cup \mathcal{I}_2 \cup \mathcal{I}_3$. Continuing this process, we can find a vector $\d \in \text{Span}\{ \U_1, \cdots, \U_{m-r}\} = \text{Range}(\U)$, such that $\mathcal{I}_{\d} = \cup_{j=1}^{m-r}\mathcal{I}_j = [m]$; meaning that $\d \in \text{Range}(\U)$ and is fully dense. Since $\d \in \text{Range}(\U)$, we have $\A^\ttt\d = \zero_r$. $\blacksquare$

\textbf{\emph{Proof of Proposition~\ref{prop:sto}:}} Consider a matrix $\A\in \R^{m\times r}$ that is tall, full rank, and incoherent, we can rewrite \eqref{eq:sto} as 
\begin{align}\label{eq:combine}
\left[ \begin{array}{cc}
\A^\ttt & \one_r
\end{array} \right] \left[ \begin{array}{c}
\d \\
-1
\end{array}\right] = \zero_{r}
\end{align}
Let us denote $\widehat{\A}^\ttt = \left[ \begin{array}{cc}
\A^\ttt & \one_r
\end{array} \right]$. Then we can see that 
\begin{enumerate*}[label={\arabic*)}]
	\item $\widehat{\A} \in \R^{(m+1)\times r}$ is tall and full rank,
	\item $\widehat{\A}$ is incoherent by Lemma \ref{lemma:incoh}.
\end{enumerate*}
We see that $\widehat{\A}$ satisfies all the conditions in Lemma \ref{lemma}, hence there exists a ${\d}\in \R^{m+1}$ such that $\widehat{\A}^\ttt{\d} = \zero_{r}$, and $\|{\d}\|_0 = m+1$. Since $\d$ is fully dense, we construct a $\widehat{\d}\in \R^{m+1}$ as
\begin{align}
\widehat{\d} := -\d / \d(m+1).
\end{align}
By this construction, we have $\widehat{\d}(m+1) = -1$.  In addition, $\widehat{\A}^\ttt\widehat{\d}=\zero_r$ as it is merely a scaled version of $\d$. Let $\overline{\d} = \widehat{\d}(1:m)\in\R^m$, then we have 
\begin{align}
\A^\ttt \overline{\d} = \one_r ,\quad \|\overline{\d}\|_0 = m.
\end{align}
Hence we managed to show the existence of a $\d$ that satisfies both \eqref{eq:sto} and \eqref{eq:fully-dense} for any $\A$ that satisfies the conditions in Proposition \ref{prop:sto}. $\blacksquare$

%

\end{document}